\documentclass[11pt]{article}

\usepackage[]{acl}

\usepackage{times}
\usepackage{latexsym}
\usepackage{amsmath}
\usepackage{amssymb}

\usepackage[T1]{fontenc}

\usepackage[utf8]{inputenc}

\usepackage{microtype}

\usepackage{inconsolata}

\usepackage{graphicx}

\title{Detecting Hallucinations for Large Language Model-based Knowledge Graph Reasoning}

\author{
  \textbf{Xinyan Zhu\textsuperscript{1}},
  \textbf{Yaoqi Liu\textsuperscript{1}},
  \textbf{Yue Gao\textsuperscript{2}},
  \textbf{Huadong Ma\textsuperscript{1}},
\\
  \textbf{Cheng Yang\textsuperscript{1}} \thanks{\ \ Corresponding Author.},
  \textbf{Chuan Shi\textsuperscript{1}} \footnotemark[1]
\\
\\
  \textsuperscript{1}Beijing University of Posts and Telecommunications, 
  \textsuperscript{2}Tsinghua University
\\
\texttt{\{zhuxinyan, yaoqiliu, mhd, yangcheng, shichuan\}@bupt.edu.cn}, \texttt{gaoyue@tsinghua.edu.cn}
}

\begin{document}
\maketitle
\begin{abstract}

Knowledge graph (KG) reasoning infers new knowledge from existing facts and is widely applied in question answering, recommendation, and decision support. With the rapid development of large language models (LLMs), LLM-based KG reasoning frameworks have become increasingly popular by leveraging retrieved KG information. However, hallucinations in LLMs remain a critical issue. Even when relevant KG knowledge is incorporated, models may still generate incorrect outputs, leading to misinformation and unreliable decisions.
Existing hallucination detection methods either focus on LLM internal states or verify consistency with retrieved contexts, but both overlook the structural information in KGs, resulting in suboptimal performance.
To address this gap, we propose \textbf{LUCID}, the first hal\textbf{LU}cination dete\textbf{C}t\textbf{I}on method for LLM-based knowle\textbf{D}ge graph reasoning frameworks. LUCID jointly leverages LLM attention scores, KG semantics, and structural information. Specifically, it extracts node and edge features from attention scores and semantic similarities, and integrates them with KG structure using a graph neural network. We also construct manually annotated benchmark datasets for evaluation. Experiments on nine datasets show that LUCID achieves state of the art performance compared to 15 baselines.
\end{abstract}

\section{Introduction}

Knowledge graph (KG) reasoning \cite{chen2020review} focuses on analyzing and deducing existing factual information in KGs to infer implicit knowledge. It supports downstream tasks such as question answering, recommendation systems, and decision-making \cite{guo2020survey, ji2021survey}.
Integrating large language models (LLMs) \cite{ouyang2022training, qwen2025qwen25technicalreport, achiam2023gpt} with KGs, particularly through LLM-based KG reasoning frameworks \cite{hu2023survey, pan2024unifying}, has become popular.
These frameworks retrieve relevant triples from KGs, incorporate them into the prompt, and guide LLMs to generate more accurate answers, enhancing performance in complex tasks.

\begin{figure}[h]
  \centering
  \includegraphics[width=\linewidth]{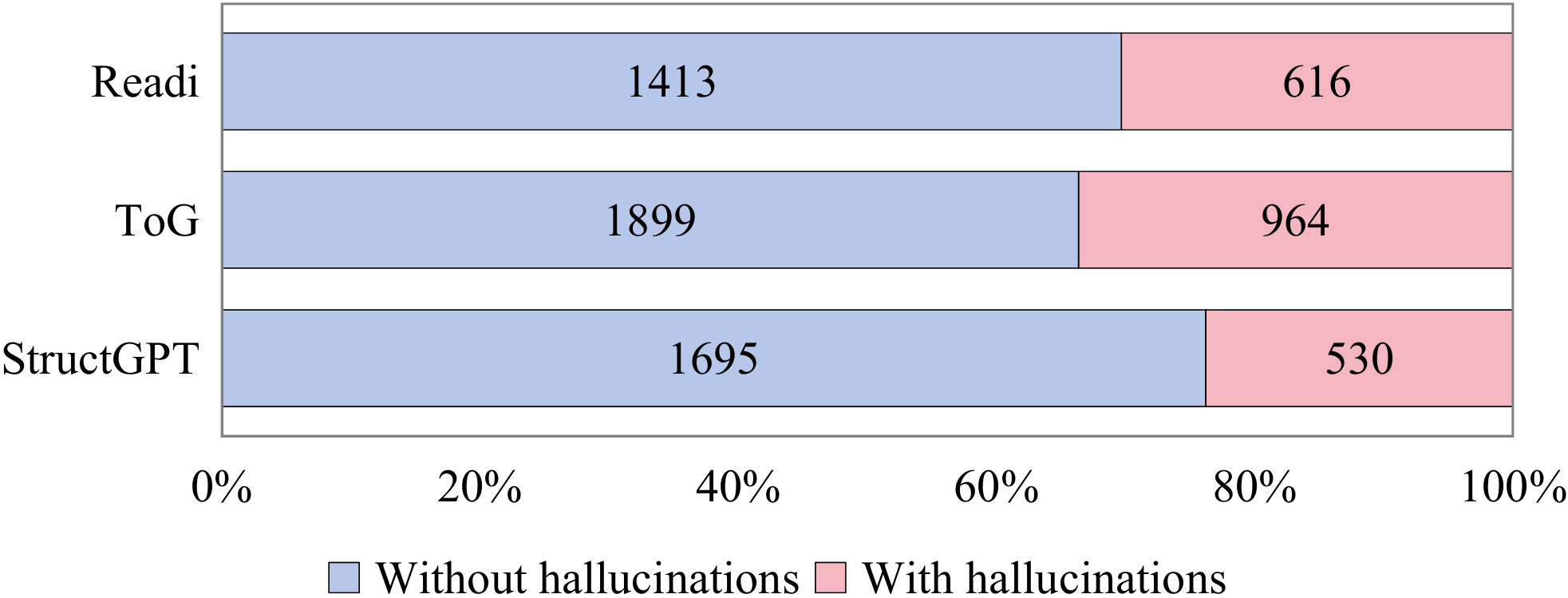}
  \caption{Distribution of responses with vs. without hallucinations across three frameworks. The marked numbers represent the count of each category.}
  \label{fig:distribution}
\end{figure}

\begin{figure}[h]
  \centering
  \includegraphics[width=\linewidth]{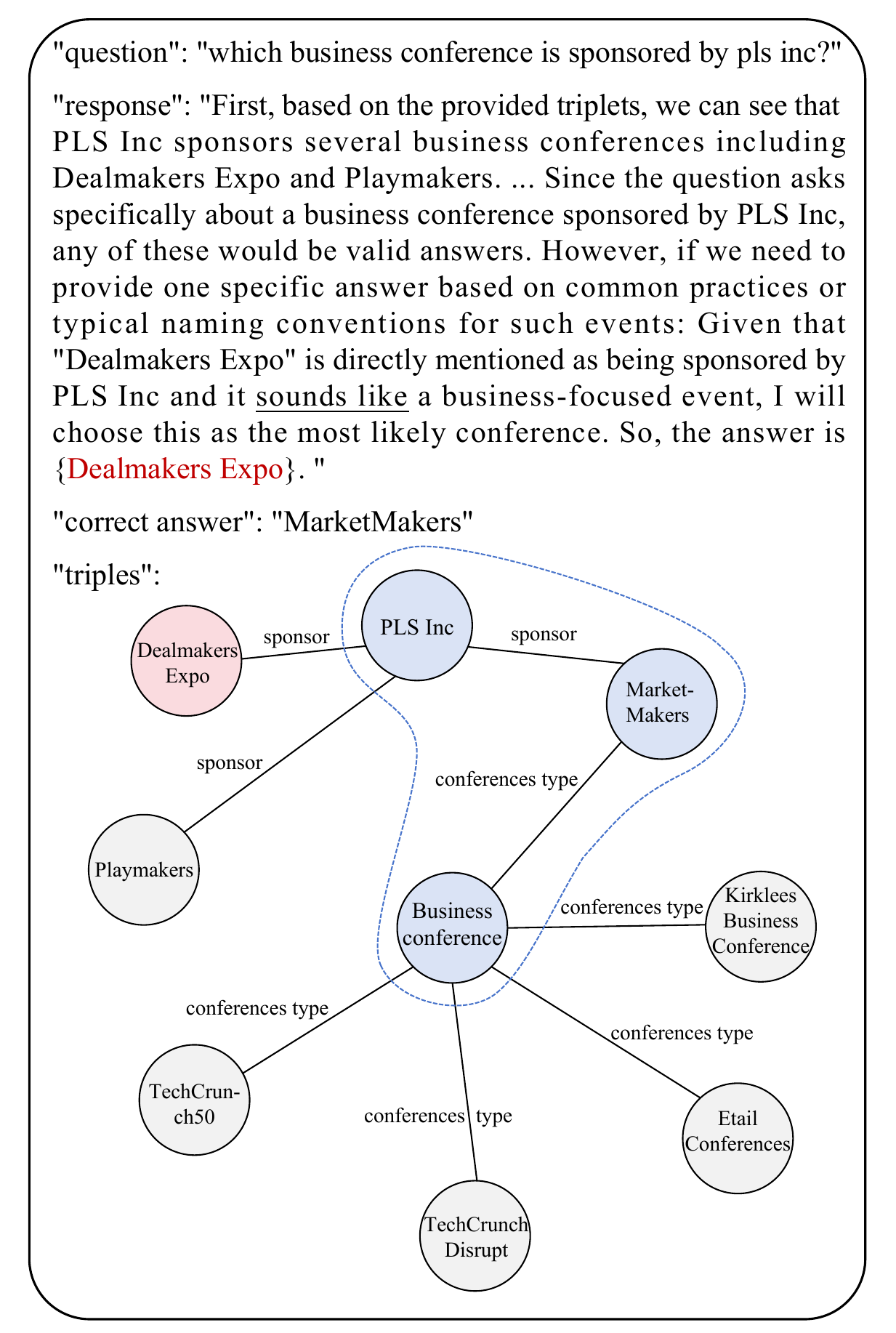}
  \caption{A hallucination example of the LLM-based KG reasoning framework, including the question, the response generated by LLMs, the correct answer, and the retrieved triples (which are visualized as a graph). The nodes and edges that can help LLMs make correct reasoning are circled in blue, while the incorrect final answers are marked in red.}
  \label{fig:sample}
\end{figure}

Although LLMs have significantly advanced KG reasoning, we find that they still produce factually misaligned responses even with retrieved knowledge, due to persistent hallucinations \cite{ji2023survey, huang2025survey}.
To quantify this, we build benchmark datasets and conduct statistical analysis by manually annotating responses from three frameworks (Readi \cite{cheng2024call}, ToG \cite{sun2023think}, StructGPT \cite{jiang2023structgpt}) on three KBQA datasets (GrailQA \cite{gu2021beyond}, WebQSP \cite{yih2016value}, QALD-10 \cite{perevalov2022qald}).
As illustrated in Figure \ref{fig:distribution}, our results reveal an average hallucination rate of 29.65\%. A representative case in Figure \ref{fig:sample} further exemplifies this problem: Although the retrieved triples clearly state that "MarketMakers" is sponsored by "PLS Inc" and belongs to the type "Business conference". LLMs incorrectly select a node that sounds like "Business conference" as the answer, causing hallucinations. 
Such hallucinations can severely compromise the accuracy of KG reasoning, leading to incorrect conclusions and misguided decision-making in downstream applications. Consequently, the development of reliable detection methods is crucial for measuring hallucinations and evaluating the inherent risks of models, providing a necessary basis for further improvements \cite{ji2023survey, li2023halueval}.

Existing methods can be applied in practical scenarios to detect hallucinations for LLM-based KG reasoning frameworks, but they still have some limitations that would result in suboptimal outcomes. General hallucination detection methods, like those using LLM internal states (e.g., EigenScore \cite{chen2024inside}), focus on the intrinsic features of models but fail to take external information into account. Even retrieval-augmented generation (RAG)-specific methods (e.g., RAGAs Faithfulness \cite{es2024ragas}) check the consistency between response and external context but ignoring the structural information provided by KGs. Effective detection requires analyzing LLM "focus" on KGs, semantic impacts of KG elements, and logical connections among them. However, no methods are tailored for the LLM-based KG reasoning frameworks.

To fill this gap, this paper proposes the first hal\textbf{LU}cination dete\textbf{C}t\textbf{I}on method for LLM-based knowle\textbf{D}ge graph reasoning frameworks (\textbf{LUCID}). 
LUCID targets a novel scenario: hallucination detection in LLM-KG integration, and uniquely integrates three types of information: (1) LLM internal state information, where the attention mechanism reflects which nodes or edges in KGs the LLMs pay more attention to during generation \cite{azaria2023internal, sun2025redeep}; (2) KG semantic information, where the semantic representation of relations determines the relevance of specific knowledge to the current task \cite{lin2015learning}; (3) KG structural information, where the graph structure explicitly represents the organizational relationships between knowledge.

We first extract attention scores of response tokens over node and edge tokens in the retrieved KG subgraph, and aggregate these scores into attention matrices across layers and heads. Then, we employ a pre-trained language model to generate embeddings for KG relations and the query, and compute their cosine similarity scores. These outputs are subsequently fused into graph features: node features are formed by flattening node attention matrices, while edge features concatenate the flattened edge attention matrices and semantic similarity scores. Finally, the subgraphs equipped with these features are fed into a trained graph neural network (GNN) model, which aggregates graph information through message passing and predicts the probability of hallucinations. We validate our method with experiments on our benchmark datasets. The results show that LUCID outperforms 15 baseline methods, achieving state-of-the-art (SOTA) performance.

The main contributions of our work are as follows: 

(1) We firstly point out the hallucinations in LLM-based KG reasoning frameworks and construct benchmark datasets through manual annotation, with the subsequent analysis uncovering an average hallucination rate of 29.65\%.

(2) We propose LUCID, the first hallucination detection method specifically designed for LLM-based KG reasoning frameworks, which integrates LLM internal state information, KG semantic and structural information.

(3) Extensive experiments demonstrate that our method outperforms all baselines, proving strong generalization. And the hallucination probability obtained by LUCID can notably help improve the accuracy of QA.

\begin{figure*}[h!]
  \centering
  \includegraphics[width=\linewidth]{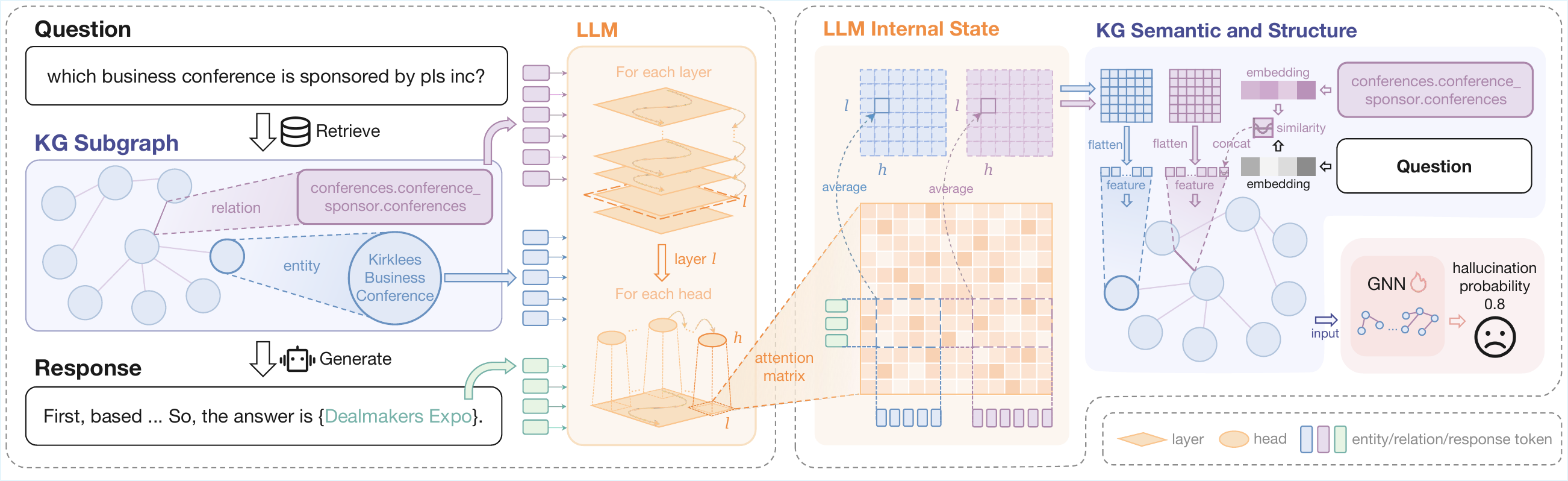}
  \caption{The architecture of LUCID. The diagram uses one representative node and edge from the graph as an example. The left portion depicts the typical workflow of an LLM-based KG reasoning framework and the relationships between the various layers and heads inside the LLMs, while the right portion shows how the method constructs a KG subgraph by integrating three information components: LLM internal state, KG semantic and structure information. This enriched subgraph is then processed by a GNN model to generate the predicted hallucination probability.}
  \label{fig:architecture}
\end{figure*}

\section{Related Work}

\subsection{General Hallucination Detection Methods}
General hallucination detection methods are extensively explored, with methods falling into three main categories: uncertainty-based, sampling-based, and LLM-based methods. 

\textbf{Uncertainty-based} methods leverage the uncertainty of LLM generations as indicators of potential hallucinations. 

LN-Entropy \cite{malinin2020uncertainty} measures sequence-level uncertainty via normalized entropy. Energy \cite{liu2020energy} detects unreliable predictions through energy functions, while Perplexity \cite{ren2022out} highlights uncertainty by assessing prediction difficulty. Focus \cite{zhang2023enhancing} mimics human factuality checks by focusing on uncertain tokens and keywords.

\textbf{Sampling-based} methods detect hallucinations by generating rewritten versions of the detected content and then measuring consistency. Examples include Lexical Similarity \cite{lin2022towards}, which compares semantic similarity across sampled responses, and SelfCheckGPT-Nli and SelfCheckGPT-BertScore \cite{manakul2023selfcheckgpt}, which use NLI and BertScore, respectively, to compare primary responses with sampled versions.

\textbf{LLM-based} methods detect hallucinations through carefully designed prompts, direct fine-tuning the model, or by examining the internal states of LLMs. EigenScore \cite{chen2024inside} measures semantic consistency by analyzing the covariance matrix of LLM embeddings. MetaQA \cite{yang2025hallucination} uses metamorphic relations and prompt mutation to detect hallucinations without external resources.

\subsection{RAG-specific Hallucination Detection Methods}
RAG-specific hallucination detection methods primarily focus on verifying the consistency between generated responses and retrieved contexts. 

RAGAs Faithfulness \cite{es2024ragas} assesses response accuracy by breaking responses into assertions and verifying them against the source context, with a score based on the ratio of supported statements. Trulens Groundedness \footnote{https://www.trulens.org/} evaluates the alignment between responses and context by measuring their overlap, producing a groundedness score. LettuceDetect \cite{kovacs2025lettucedetect}, a token classification method based on ModernBERT and trained on the RAGTruth \cite{wu2024ragtruth} dataset, detects hallucinated tokens in context-question-answer triples. ReDeEP \cite{sun2025redeep} predicts hallucinations by regressing token-level or chunk-level External Context and Parametric Knowledge Scores, which are highly correlated with hallucination labels.

\subsection{LLM-based KG Reasoning Frameworks}\label{framework}

LLM-based KG reasoning frameworks leverage LLMs to overcome the limitations of traditional KGs in complex reasoning. By integrating structured KGs with retrieved textual evidence, these frameworks enable LLMs to reason over KG structure, supporting tasks such as multi-hop reasoning and question answering (QA).

Different strategies are used to integrate KGs into reasoning. Readi (Reasoning Path Editing) \cite{cheng2024call} allows LLMs to generate reasoning paths for queries and instantiate them with KGs, triggering path editing when errors occur. ToG (Think-on-Graph) \cite{sun2023think} enables LLMs to iteratively search KGs, refining paths to gather sufficient information for answers. StructGPT \cite{jiang2023structgpt} introduces an Iterative Reading-then-Reasoning method, where LLMs interact with specialized KG interfaces to extract structured information and refine reasoning through generation cycles.

\section{Methodology}\label{Methodology}

\subsection{Overview}

We propose to detect hallucinations for LLM-based KG reasoning by fusing LLM internal state, KG semantic and structural information. Specifically, our method LUCID first processes the LLM internal states to generate node and edge attention matrices. It then computes semantic similarities between the KG relations and the query, combining these elements to form comprehensive node and edge features. These integrated features are subsequently used to train a GNN for hallucination probability prediction. The architecture of LUCID can be seen in Figure \ref{fig:architecture}. 

\subsection{Problem Statement}

\textbf{LLM-based KG Reasoning Frameworks}. In processing a user query $Q$, LLM-based KG reasoning frameworks first retrieve relevant information from KGs to form a subgraph, denoted with nodes $V$ (entities) and edges $E$ (relations). This subgraph is then serialized into text $S_Q$, which is then concatenated with $Q$ and fed into the LLMs to generate final response $R_Q$.

\textbf{Hallucinations in LLM-based KG Reasoning}.

Hallucinations in LLMs \cite{huang2025survey} refer to the generation of content that is inconsistent with objective facts, real data, or the given context. In LLM-based KG reasoning frameworks, hallucinations follow the same definition: despite access to retrieved knowledge, the model may still produce statements unsupported by factual or contextual evidence \cite{wu2024ragtruth}. We therefore formally define hallucinations as follows:

If any statement in the response $R_Q$ is inconsistent with the objective facts or the information in $S_Q$, then $R_Q$ has hallucinations, otherwise it hasn't.

\textbf{Task Definition}.
We denote the hallucination detection task in LLM-based KG reasoning frameworks as follows:

Given a query \( Q \), a retrieved KG subgraph text $S_Q$, and an LLM-generated response \( R_Q \) based on \( S_Q \), the task aims to learn a detection function \( \mathcal{D} \) that outputs whether \( R_Q \) has hallucinations based on available information:
\[ 
\quad \mathcal{D}(Q, S_Q, R_Q) = 
\begin{cases}
1, & \text{if } R_Q \text{ has hallucinations} , \\
0, & \text{otherwise}.
\end{cases}
\]

\subsection{LLM Internal State Information Processing}

The internal states of LLMs can provide various types of information, among which the attention mechanism plays a crucial role in enabling context understanding and modeling token relationships \cite{vaswani2017attention}. To take advantage of this capability, we specifically analyze the attention scores calculated by LLMs when generating response tokens.

Within the Transformer architecture, attention is computed through a structure composed of multiple layers and heads, which directly governs the ability of models to comprehend, associate, and reason over input information. A "layer" refers to a hierarchical stage in the model responsible for progressively extracting and transforming representations of the input. Each "layer" contains multiple parallel "heads", which are individual attention calculation units in the multi-head self-attention mechanism, used to capture the dependencies between tokens in the sequence from different perspectives \cite{clark2019does}.


Formally, given an input context $C = Q \parallel S_Q$, where the symbol $\parallel$ denotes concatenation, and $S_Q$ represents the textually serialized subgraph retrieved by LLMs based on query $Q$. We define $T_{S_Q} = \{t_1, t_2, ..., t_i\}$ as tokens in $C$ corresponding to the elements in $S_Q$, where $i = \lvert T_{S_Q} \lvert$. During response generation, the model first generates intermediate reasoning steps before outputting the final response tokens $A$. For each token $a \in A$, LLMs calculate attention scores $\alpha_{a,t} \in [0,1]$ over all tokens $t \in T_{S_Q}$, representing its contextual focus. For ease of description, we uniformly denote the nodes (entities) in KGs by $e$ and the edges (relations) by $r$.

To aggregate these scores at the node and edge levels, we compute the average attention score at each layer, per head, for all response tokens to all nodes or edge tokens. For a node $e$ in $S_Q$, let $T_e \subseteq T_{S_Q}$ be the tokens corresponding to $e$. The attention score for node $e$ in layer $l$ and head $h$ is:
\begin{equation}
    \alpha_{e,l,h} = \frac{1}{|T_e| \cdot |A|} \sum_{a \in A} \sum_{t \in T_e} \alpha_{a,t,l,h},
\end{equation}
where $\alpha_{a,t,l,h}$ is the attention score in layer $l$, head $h$, corresponding to the response token $a$ and the node token $t$.

Similarly, for an edge $r$ in $S_Q$, let $T_r \subseteq T_{S_Q}$ be the tokens corresponding to $r$. The attention score for edge $r$ in layer $l$ and head $h$ is:

\begin{equation}
    \alpha_{r,l,h} = \frac{1}{|T_r| \cdot |A|} \sum_{a \in A} \sum_{t \in T_r} \alpha_{a,t,l,h}.
\end{equation}

For each layer $l \in \{1, ..., L\}$ and head $h \in \{1, ..., H\}$, we compute attention scores over the KG subgraphs, aggregating them into two mean attention matrices: $\mathbf{M}_e \in \mathbb{R}^{L \times H}$ for nodes, $\mathbf{M}_r \in \mathbb{R}^{L \times H}$ for edges.

\subsection{KG Semantic and Structural Information Processing}

The relevance of KG elements to the query directly affects the reliability of reasoning, and semantic similarity can quantify this relevance. Therefore, we leverage the semantic similarity between each edge in the KG subgraph and the question to distinguish the "task-related" from "task-irrelevant" edges. Nodes correspond to KG entities, whose surface forms are often ambiguous and provide limited semantic signal for query relevance. As a result, semantic relevance is computed only for relations.

For each edge in the retrieved KG subgraph, we first obtain the textual description of its relations. Then we employ the lightweight Transformer-based pre-trained language model all-MiniLM-L6-v2\footnote{https://huggingface.co/sentence-transformers/all-MiniLM-L6-v2} to generate embeddings for both relations and queries. After obtaining the relation embedding $\text{emb}(r) \in \mathbb{R}^d$ and the query embedding $\text{emb}(Q) \in \mathbb{R}^d$ (where $d$ denotes the dimension of the output features), we compute the semantic similarity score $s(r, Q)$ for relation $r$ and query $Q$ using cosine similarity.

Instead of directly using the triple literals of the KG subgraphs for hallucination detection, we construct a graph from the triples and build features for its edges and nodes based on the information obtained from the above two processing. Given that $V$ denotes the set of nodes and $E$ denotes the set of edges, the constructed subgraph is represented as $G = (V, E)$. 

Once the subgraph is constructed, the initial node and edge features are formed by flattening their corresponding attention matrices into one-dimensional vectors. Specifically, for a node $e$, its feature is defined as \(\mathbf{x}_e = \text{flatten}(\mathbf{M}_e)\). For an edge $r$, its feature is enhanced by concatenating semantic similarity, formulated as \(\mathbf{x}_r = \text{concat}( \text{flatten}(\mathbf{M}_r), s(r, Q))\).

\begin{table*}[t]
\centering
\fontsize{8.6pt}{10pt}\selectfont
\setlength{\tabcolsep}{2pt}   
\renewcommand{\arraystretch}{1.3}
\begin{tabular}{l | c c c c | c c c c | c c c c}
\hline
\textbf{Method} & \multicolumn{4}{c}{\textbf{GrailQA}} & \multicolumn{4}{c}{\textbf{WebQSP}} & \multicolumn{4}{c}{\textbf{QALD-10}} \\
\hline
 & ACC & AUC & PCC & AVG & ACC & AUC & PCC & AVG & ACC & AUC & PCC & AVG \\
\hline
LLM & \underline{0.7705} & 0.6301 & 0.3545 & 0.5850 & \underline{0.7299} & 0.6289 & 0.2970 & 0.5519 & 0.6923 & 0.6779 & \underline{0.3480} & 0.5727 \\
Perplexity & 0.6410 & 0.7150 & 0.3160 & 0.5573 & 0.6299 & 0.6688 & 0.2368 & 0.5118 & 0.6390 & 0.6624 & 0.2462 & 0.5159 \\
Energy & 0.5254 & 0.5543 & 0.0368 & 0.3722 & 0.4729 & 0.4555 & -0.0549 & 0.2912 & 0.5024 & 0.4416 & -0.0186 & 0.3085 \\
LN-Entropy & 0.6933 & 0.7315 & 0.3168 & 0.5805 & 0.6337 & 0.6581 & 0.2268 & 0.5062 & 0.6000 & 0.6582 & 0.2117 & 0.4900 \\
Lexical Similarity & 0.6479 & 0.7036 & 0.2782 & 0.5432 & 0.5861 & 0.5746 & 0.1159 & 0.4255 & 0.6098 & 0.6299 & 0.1898 & 0.4765 \\
SelfCk-Nli & 0.6575 & 0.7527 & 0.3787 & 0.5963 & 0.6765 & 0.6954 & 0.3353 & 0.5691 & 0.6878 & 0.7246 & 0.3469 & 0.5864 \\
SelfCk-BERTScore & 0.6176 & 0.6775 & 0.2186 & 0.5046 & 0.6089 & 0.6477 & 0.2436 & 0.5001 & 0.6098 & 0.6691 & 0.2934 & 0.5241 \\
Focus & 0.5395 & 0.5253 & 0.0475 & 0.3708 & 0.5395 & 0.5253 & 0.0778 & 0.3809 & 0.5610 & 0.4837 & 0.0618 & 0.3688 \\
EigenScore & 0.6052 & 0.6327 & 0.2051 & 0.4810 & 0.5005 & 0.5326 & 0.0726 & 0.3686 & 0.5463 & 0.5447 & 0.1412 & 0.4107 \\
MetaQA & 0.5392 & 0.5522 & 0.1043 & 0.3986 & 0.6175 & 0.5578 & 0.1121 & 0.4291 & 0.6000 & 0.6558 & 0.2117 & 0.4892 \\
\hline
RAGAs & 0.7015 & 0.7341 & 0.3578 & 0.5978 & 0.6394 & 0.6949 & 0.3080 & 0.5474 & 0.5610 & 0.6179 & 0.2013 & 0.4601 \\
Trulens & 0.7538 & \textbf{0.8202} & 0.4800 & \underline{0.6847} & 0.6969 & \underline{0.7636} & \underline{0.4010} & \underline{0.6205} & 0.6404 & 0.6778 & 0.2630 & 0.5271 \\
LettuceDetect & 0.7455 & 0.5222 & \underline{0.4611} & 0.5763 & 0.6108 & 0.4399 & 0.3337 & 0.4615 & 0.4829 & 0.4000 & 0.2724 & 0.3851 \\
ReDeEP (token) & 0.7125 & 0.7850 & 0.3931 & 0.6302 & 0.6546 & 0.6487 & 0.2481 & 0.5171 & 0.6098 & 0.6583 & 0.3072 & 0.5251 \\
ReDeEP (chunk) & 0.7262 & 0.7914 & 0.4122 & 0.6433 & 0.6841 & 0.7488 & 0.3125 & 0.5818 & \underline{0.6976} & \underline{0.7529} & 0.3466 & \underline{0.5990} \\
\hline
LUCID & \textbf{0.7882*} & \underline{0.8045} & \textbf{0.5171*} & \textbf{0.7033*} & \textbf{0.7307} & \textbf{0.7910*} & \textbf{0.4043} & \textbf{0.6420*} & \textbf{0.7171*} & \textbf{0.7616*} & \textbf{0.3629*} & \textbf{0.6139*} \\
\hline
\end{tabular}
\caption{\label{tab:main-readi}
Main results of the hallucination detection performance of LUCID and 15 baseline methods on three datasets GrailQA, WebQSP and QALD-10 under the Readi framework. The evaluation metrics include ACC, AUC, PCC and the average of the three (AVG). Bold and underline fonts denote the best and second-best among all methods. * indicates that LUCID significantly outperforms the optimal baseline at the 99\% significance level.
}
\end{table*}

\begin{table*}[t]
\centering
\fontsize{8.6pt}{10pt}\selectfont
\setlength{\tabcolsep}{2pt}   
{
\renewcommand{\arraystretch}{1.3}
\begin{tabular}{l | c c c c | c c c c | c c c c}
\hline
\textbf{Method} & \multicolumn{4}{c}{\textbf{GrailQA}} & \multicolumn{4}{c}{\textbf{WebQSP}} & \multicolumn{4}{c}{\textbf{QALD-10}} \\
\hline
 & ACC & AUC & PCC & AVG & ACC & AUC & PCC & AVG & ACC & AUC & PCC & AVG \\
\hline
LLM & 0.7018 & 0.6026 & 0.2806 & 0.5283 & 0.6637 & 0.6518 & 0.3306 & 0.5487 & \underline{0.7089} & 0.6697 & 0.3837 & 0.5874 \\
Perplexity & 0.6704 & 0.7091 & 0.2368 & 0.5388 & 0.6228 & 0.6310 & 0.1940 & 0.4826 & 0.6174 & 0.6421 & 0.2500 & 0.5032 \\
Energy & 0.4439 & 0.3830 & -0.0549 & 0.2573 & 0.3928 & 0.3845 & -0.1765 & 0.2003 & 0.5370 & 0.5135 & 0.0748 & 0.3751 \\
LN-Entropy & 0.6480 & 0.6772 & 0.2268 & 0.5173 & 0.6020 & 0.6337 & 0.1867 & 0.4741 & 0.6431 & 0.6492 & 0.2884 & 0.5269 \\
Lexical Similarity & 0.6054 & 0.6590 & 0.1159 & 0.4601 & 0.5772 & 0.5900 & 0.1377 & 0.4350 & 0.5820 & 0.5963 & 0.1571 & 0.4451 \\
SelfCk-Nli & 0.6297 & 0.7166 & 0.3119 & 0.5527 & \underline{0.6758} & 0.7143 & \underline{0.3753} & \underline{0.5885} & 0.6603 & 0.6964 & 0.3439 & 0.5669 \\
SelfCk-BERTScore & 0.7115 & \underline{0.7596} & 0.3228 & \underline{0.5980} & 0.6452 & 0.6818 & 0.3002 & 0.5424 & 0.6635 & 0.7246 & 0.3792 & 0.5891 \\
Focus & 0.6491 & 0.5476 & 0.1335 & 0.4434 & 0.5065 & 0.5166 & 0.0748 & 0.3660 & 0.5673 & 0.5615 & 0.1353 & 0.4214 \\
EigenScore & 0.5191 & 0.5814 & 0.1264 & 0.4090 & 0.5303 & 0.5351 & 0.0405 & 0.3686 & 0.5466 & 0.5451 & 0.0951 & 0.3956 \\
MetaQA & 0.4790 & 0.5803 & 0.0867 & 0.3820 & 0.4707 & 0.5397 & 0.0133 & 0.3412 & 0.4968 & 0.4969 & 0.0052 & 0.3330 \\
\hline
RAGAs & 0.6588 & 0.6839 & 0.2795 & 0.5407 & 0.6087 & 0.6589 & 0.2602 & 0.5093 & 0.6506 & 0.6870 & 0.3536 & 0.5637 \\
Trulens & 0.6573 & 0.7245 & 0.3113 & 0.5644 & 0.6562 & 0.6769 & 0.2826 & 0.5386 & 0.6977 & 0.7340 & 0.4146 & 0.6154 \\
LettuceDetect & 0.6868 & 0.4582 & \underline{0.3239} & 0.4896 & 0.5872 & 0.4481 & 0.2792 & 0.4382 & 0.4739 & 0.4034 & 0.4121 & 0.4298 \\
ReDeEP (token) & \underline{0.7191} & 0.7355 & 0.3103 & 0.5883 & 0.6699 & 0.7144 & 0.2433 & 0.5425 & 0.6731 & 0.6802 & 0.3790 & 0.5774 \\
ReDeEP (chunk) & 0.7061 & 0.7540 & 0.3232 & 0.5944 & 0.6367 & \underline{0.7185} & 0.3580 & 0.5711 & 0.7019 & \underline{0.7636} & \underline{0.4147} & \underline{0.6267} \\
\hline
LUCID & \textbf{0.7287*} & \textbf{0.7712*} & \textbf{0.3316*} & \textbf{0.6105*} & \textbf{0.6882*} & \textbf{0.7368*} & \textbf{0.3790} & \textbf{0.6013*} & \textbf{0.7340*} & \textbf{0.7832*} & \textbf{0.4302*} & \textbf{0.6491*} \\
\hline
\end{tabular}
}
\caption{\label{tab:main-ToG}
Main results of the hallucination detection performance of LUCID and 15 baseline methods on three datasets GrailQA, WebQSP and QALD-10 under the ToG framework. The evaluation metrics include ACC, AUC, PCC and the average of the three (AVG). Bold and underline fonts denote the best and second-best among all methods. * indicates that LUCID significantly outperforms the optimal baseline at the 99\% significance level.
}
\end{table*}

\subsection{GNN for Hallucination Detection}

After the subgraph structure $G = (V, E)$ is prepared and its nodes and edges are associated with their respective features $\{\mathbf{x}_e\}_{e \in V}$ and $\{\mathbf{x}_r\}_{r \in E}$, we use graph isomorphism network with edge features (GINE) \cite{hu2019strategies} model trained to predict the probability of hallucinations. The GINE model extends graph isomorphism networks (GIN) \cite{xu2018how} by explicitly incorporating edge features into message passing. This enhancement allows GINE to leverage rich edge information, thereby achieving more expressive power than GIN when working with graphs that have meaningful edge attributes. As a result, when applied to our task, the trained GINE model can effectively identify hallucinations in the outputs of other LLM-based KG reasoning frameworks.

Formally, for each node $e$, its feature is iteratively updated across $K$ layers using information from its neighbors. In layer $k$, the updated feature $\mathbf{h}_e^{(k)}$ is computed as: 

\begin{equation}
\begin{split}
    \mathbf{h}_e^{(k)} &= \text{MLP}^{(k)} \bigg( \mathbf{h}_e^{(k-1)} \\ 
    & + \sum_{(u, e) \in E} \text{ReLU}\left( \mathbf{h}_u^{(k-1)} + \mathbf{x}_{(u,e)} \right) \bigg),
\end{split}
\end{equation}
where $\mathbf{h}_e^{(0)} = \mathbf{x}_e$, $\text{MLP}^{(k)}$ is a multilayer perceptron for layer $k$, and $\mathbf{x}_{(u,e)}$ is the edge features between node $u$ and $e$.
\nobreak
After $K$ layers, a graph-level representation is obtained by pooling node features: $\mathbf{h}_G = \text{sum}\left( \{\mathbf{h}_e^{(K)}\}_{e \in V} \right)$. This graph representation is fed into a final classifier to predict the hallucination probability $p = \sigma\left( \mathbf{W} \mathbf{h}_G + \textbf{b} \right)$, where $\sigma$ is the sigmoid function, and $\mathbf{W}, \textbf{b}$  are learnable parameters.

Based on manually annotated datasets, we use label 1 for responses with hallucinations and label 0 for responses without hallucinations. The model is then trained to perform binary classification by minimizing the binary cross-entropy loss.

To determine the detection threshold, we employ the method of maximizing the geometric mean based on the receiver operating characteristic (ROC).
Specifically, we compute the ROC curve by evaluating the true positive rate (TPR) and false positive rate (FPR) under different classification thresholds.
The threshold $\tau^*$ is selected to maximize the geometric mean of sensitivity (TPR) and specificity (1 - \text{FPR}) \cite{davis2006relationship}:

\begin{equation}
 \tau^* = \mathop{\mathrm{argmax}}_{\tau} \sqrt{\text{TPR}(\tau) \cdot (1 - \text{FPR}(\tau))}.
\end{equation}

\section{Experiments}

\begin{table*}[t]
\centering
\fontsize{8.6pt}{10pt}\selectfont
\setlength{\tabcolsep}{2pt}
{
\renewcommand{\arraystretch}{1.3}
\begin{tabular}{l | c c c c | c c c c | c c c c}
\hline
\textbf{Method} & \multicolumn{4}{c}{\textbf{GrailQA}} & \multicolumn{4}{c}{\textbf{WebQSP}} & \multicolumn{4}{c}{\textbf{QALD-10}} \\
\hline
 & ACC & AUC & PCC & AVG & ACC & AUC & PCC & AVG & ACC & AUC & PCC & AVG \\
\hline
LLM & 0.7163 & 0.5729 & 0.2263 & 0.5051 & \underline{0.6398} & \textbf{0.7263} & 0.1803 & 0.5155 & 0.7433 & 0.5987 & 0.3159 & 0.5526 \\
Perplexity & 0.6754 & 0.7613 & 0.3921 & 0.6096 & 0.6175 & 0.6314 & 0.1851 & 0.4780 & 0.6866 & 0.7364 & 0.3805 & 0.6012 \\
Energy & 0.3208 & 0.5781 & 0.0419 & 0.3136 & 0.4437 & 0.4936 & 0.0070 & 0.3148 & 0.4963 & 0.4658 & -0.0411 & 0.3070 \\
LN-Entropy & 0.7080 & 0.7626 & 0.4053 & 0.6253 & 0.5675 & 0.6554 & 0.2063 & 0.4764 & 0.6866 & 0.7204 & 0.3417 & 0.5829 \\
Lexical Similarity & 0.6579 & 0.7326 & 0.3423 & 0.5776 & 0.5939 & 0.6191 & 0.1588 & 0.4573 & 0.6903 & 0.6936 & 0.3344 & 0.5728 \\
SelfCheckGPT-Nli & 0.7252 & 0.7863 & 0.4365 & \underline{0.6493} & 0.6245 & 0.7062 & \textbf{0.2464} & \underline{0.5257} & 0.7201 & \textbf{0.8129} & \underline{0.4218} & 0.6516 \\
SelfCheckGPT-BERTScore & 0.7030 & \underline{0.7952} & \underline{0.4477} & 0.6486 & 0.5869 & 0.6544 & 0.2275 & 0.4896 & 0.7015 & 0.7527 & 0.4106 & 0.6216 \\
Focus & 0.5451 & 0.4627 & 0.0155 & 0.3411 & 0.2197 & 0.4564 & -0.0180 & 0.2194 & 0.6455 & 0.5432 & 0.1599 & 0.4495 \\
EigenScore & 0.5539 & 0.6220 & 0.1855 & 0.4538 & 0.5758 & 0.6042 & 0.1618 & 0.4473 & 0.6231 & 0.6378 & 0.2211 & 0.4940 \\
MetaQA & 0.5727 & 0.5764 & 0.1396 & 0.4296 & 0.5216 & 0.5691 & 0.0740 & 0.3882 & 0.5858 & 0.6293 & 0.2598 & 0.4916 \\
\hline
RAGAs Faithfulness & 0.6688 & 0.6989 & 0.3210 & 0.5629 & 0.5633 & 0.6253 & 0.2255 & 0.4714 & 0.5933 & 0.6584 & 0.2741 & 0.5086 \\
Trulens Groundedness & 0.6876 & 0.7410 & 0.3741 & 0.6009 & 0.6134 & 0.6854 & 0.1683 & 0.4890 & 0.6217 & 0.7225 & 0.3973 & 0.5805 \\
LettuceDetect & 0.5301 & 0.3891 & 0.2306 & 0.3833 & 0.3561 & 0.2416 & 0.1212 & 0.2396 & 0.6731 & 0.6035 & 0.2434 & 0.5067 \\
ReDeEP (token) & 0.6867 & 0.7527 & 0.3203 & 0.5866 & 0.4840 & 0.5262 & 0.1943 & 0.4015 & 0.6045 & 0.6463 & 0.3989 & 0.5499 \\
ReDeEP (chunk) & \textbf{0.7419} & 0.7865 & 0.4164 & 0.6483 & 0.5522 & 0.6134 & 0.2047 & 0.4568 & \underline{0.7463} & 0.8027 & 0.4178 & \underline{0.6556} \\
\hline
LUCID & \underline{0.7331} & \textbf{0.7958} & \textbf{0.4485} & \textbf{0.6591*} & \textbf{0.6634*} & \underline{0.7206} & \underline{0.2292} & \textbf{0.5377*} & \textbf{0.7537*} & \underline{0.8122} & \textbf{0.4255} & \textbf{0.6638*} \\
\hline
\end{tabular}
}
\caption{\label{tab:main-StructGPT}
Main results of the hallucination detection performance of LUCID and 15 baseline methods on three datasets GrailQA, WebQSP and QALD-10 under the StructGPT framework. The evaluation metrics include ACC, AUC, PCC and the average of the three (AVG). Bold and underline fonts denote the best and second-best among all methods. * indicates that LUCID significantly outperforms the optimal baseline at the 99\% significance level.
}
\end{table*}

\begin{table*}[t]
\centering
\fontsize{7.5pt}{10pt}\selectfont
\setlength{\tabcolsep}{1.5pt}
{
\renewcommand{\arraystretch}{1.3}
\begin{tabular}{l | c c c c c c c c c c c c c c c c c c}
\hline
 & \multicolumn{6}{c}{\textbf{Readi}} & \multicolumn{6}{c}{\textbf{ToG}} & \multicolumn{6}{c}{\textbf{StructGPT}} \\
\hline
 & \multicolumn{2}{c}{\textbf{GrailQA}} & \multicolumn{2}{c}{\textbf{WebQSP}} & \multicolumn{2}{c}{\textbf{QALD-10}} 
 & \multicolumn{2}{c}{\textbf{GrailQA}} & \multicolumn{2}{c}{\textbf{WebQSP}} & \multicolumn{2}{c}{\textbf{QALD-10}} 
 & \multicolumn{2}{c}{\textbf{GrailQA}} & \multicolumn{2}{c}{\textbf{WebQSP}} & \multicolumn{2}{c}{\textbf{QALD-10}} \\
\hline
 & EM & Cost & EM & Cost & EM & Cost 
 & EM & Cost & EM & Cost & EM & Cost 
 & EM & Cost & EM & Cost & EM & Cost \\
\hline
Qwen2.5-7B & 69.18 & - & 69.83 & - & 50.24 & - & 60.47 & - & 67.25 & - & 46.43 & - & 61.02 & - & 71.76 & - & 54.10 & - \\
Qwen3-235B & 77.03 & 31.13 & 75.17 & 58.45 & 62.93 & 7.13 & 67.91 & 30.86 & 74.27 & 64.25 & 61.92 & 11.02 & 68.17 & 23.57 & 76.49 & 22.84 & 60.45 & 7.85 \\
Mixed & 76.34 & 11.00 & 74.31 & 25.97 & 62.93 & 3.06 & 66.67 & 14.11 & 73.20 & 26.77 & 60.06 & 5.15 & 67.79 & 10.04 & 75.80 & 11.79 & 59.33 & 3.95 \\
\hline
$\Delta$ & -0.9\% & -64.7\% & -1.1\% & -55.6\% & 0.0\% & -57.1\% 
& -0.8\% & -54.3\% & -1.4\% & -58.3\% & -3.0\% & -53.2\% 
& -0.6\% & -57.4\% & -0.9\% & -48.4\% & -1.9\% & -49.7\% \\
\hline
\end{tabular}
}
\caption{\label{tab:QA}
Comparison of QA strategies with and without LUCID-guided refinement.
The EM column reports accuracy in percentage, whereas the Cost column is measured in dollars (\$).
The first two strategies represent using Qwen2.5-7B and Qwen3-235B for QA tasks on 9 datasets.
Mixed refers to the process where, when LUCID detects a high hallucination probability in Qwen2.5-7B outputs, these cases are reprocessed by Qwen3-235B.
The $\Delta$ row represents the percentage difference between the Mixed and Qwen3-235B model.
}
\end{table*}

\subsection{Experimental Setting}
\subsubsection{Datasets}
\label{dataset}
The KBQA datasets used in the experiment are as follows:

\textbf{CWQ} \cite{talmor2018web} is a dataset that features complex, broad questions requiring reasoning over multiple information pieces and diverse paraphrasing, designed to test models' ability to decompose questions and integrate answers for complex QA tasks.

\textbf{WebQSP} \cite{yih2016value} is a single-hop and multi-hop QA dataset with annotated SPARQL queries, commonly used to evaluate semantic parsing and structured reasoning on complex questions.

\textbf{GrailQA} \cite{gu2021beyond} is a large-scale multilingual question answering dataset that covers complex scenarios including multi-hop reasoning and aggregate queries. It supports evaluation across three generalization levels to assess model universality and robustness.

\textbf{QALD-10} \cite{perevalov2022qald} is a multilingual KGQA benchmark that features complex questions. It primarily evaluates model performance in cross-lingual and cross-domain reasoning tasks. In our experiments we selected the English subset of this dataset.

\subsubsection{Benchmark}

In order to evaluate whether the methods can effectively detect hallucinations in the LLM-based KG reasoning frameworks, we label the responses generated by the three frameworks (Readi \cite{cheng2024call}, ToG \cite{sun2023think}, StructGPT \cite{jiang2023structgpt}) from KBQA datasets through manual annotation.
A response is labeled as hallucinated if it contains at least one factual statement that is not supported by, or contradicts, objective facts or the retrieved KG subgraphs. Factual validity is verified against authoritative encyclopedic sources (e.g., Wikidata) when necessary. Two annotators independently perform hallucination annotation, and inter-annotator agreement is measured using Cohen’s kappa. Disagreements are resolved by a third expert. Minor discrepancies in the number of annotated samples arise from excluding invalid responses, excessively large subgraphs, or cases where attention extraction fails due to out-of-memory (OOM) errors.

For GINE model training, we adopt the COMPLEXWEBQUESTIONS (CWQ) \cite{talmor2018web} dataset to construct the training data. For testing the effectiveness of our method, we use three datasets (WebQSP \cite{yih2016value}, GrailQA \cite{gu2021beyond}, QALD-10 \cite{perevalov2022qald}) to build benchmark datasets. 

\subsubsection{Baselines}

We compare LUCID with 15 baselines, which can be divided into two categories. One is general hallucination detection methods: LLM (Leveraging the LLM's own capabilities for detection), LN-Entropy \cite{malinin2020uncertainty}, Energy \cite{liu2020energy}, Perplexity \cite{ren2022out}, Lexical Similarity \cite{lin2022towards}, SelfCheckGPT-Nli \cite{manakul2023selfcheckgpt}, SelfCheckGPT-BERTScore \cite{manakul2023selfcheckgpt}, Focus \cite{zhang2023enhancing}, EigenScore \cite{chen2024inside}, MetaQA \cite{yang2025hallucination}; the other is RAG-specific hallucination detection methods: RAGAs Faithfulness \cite{es2024ragas}, Trulens Groundedness, LettuceDetect \cite{kovacs2025lettucedetect}, ReDeEP (token) \cite{sun2025redeep}, ReDeEP (chunk) \cite{sun2025redeep}. 

For methods that require open-source LLMs to be deployed, we use the Qwen2.5-7B-Instruct model; for methods that need to use closed-source LLM, we use the GPT-4o-mini model.

\subsubsection{Model Setup}
We use the subgraphs and responses generated by the Readi framework on the CWQ dataset to train GNN. For model parameters, we use two layers of GINEConv, set the number of channels in the hidden layer to 512, set the learning rate to $1\times10^{-3}$, and train 300 epochs.

\subsubsection{Evaluation Metrics} 
In order to comprehensively evaluate the performance of all methods in hallucination detection, we use the evaluation metrics: ACC (Accuracy), AUC (Area Under the ROC Curve), PCC (Pearson Correlation Coefficient), and AVG, where AVG is the average of ACC, AUC, and PCC.

\begin{table}[t]
\centering
\small
\setlength{\tabcolsep}{4.5pt}
{
\renewcommand{\arraystretch}{1.3}
\begin{tabular}{l | c c c | c}
\hline
\textbf{} & \textbf{GrailQA} & \textbf{WebQSP} & \textbf{QALD-10} & \textbf{Avg} \\
\hline
Readi & 66.85\% & 87.25\% & 68.77\% & 74.29\% \\
ToG & 81.49\% & 93.16\% & 98.45\% & 91.03\% \\
StructGPT & 98.12\% & 99.30\% & 93.68\% & 97.03\% \\
\hline
\end{tabular}
}
\caption{\label{tab:compare}
Percentage of sparse graphs among KG subgraphs retrieved by three frameworks on three datasets. The last column reports the average percentage across datasets for each framework.
}
\end{table}

\begin{figure*}[h!]
  \centering
  \includegraphics[width=\linewidth]{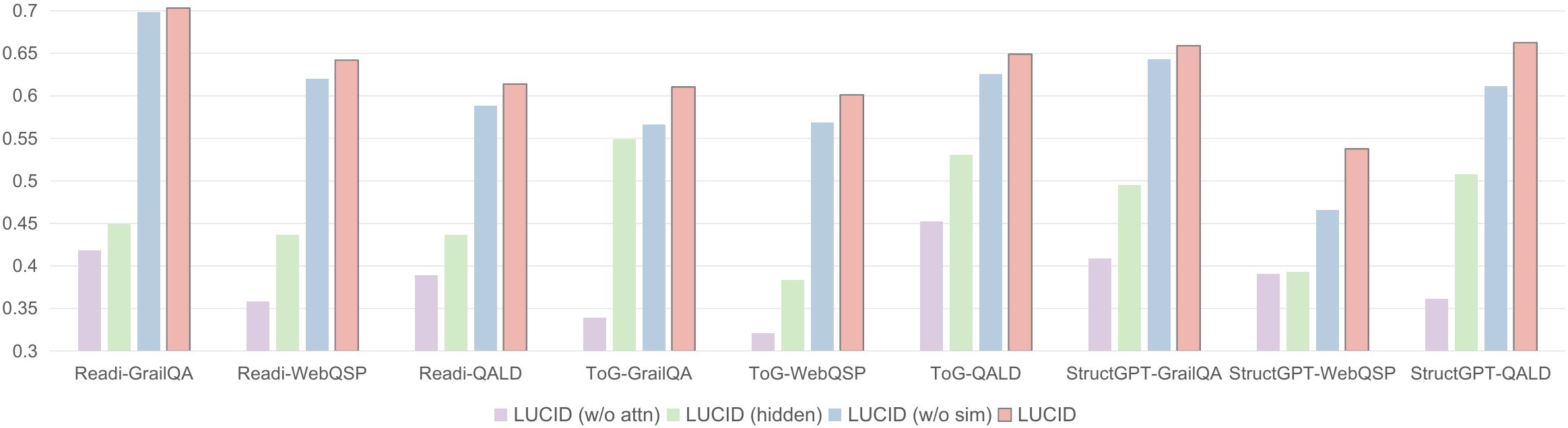}
  \caption{AVG performance of the method on three frameworks and three datasets when using different features.}
  \label{fig:ablation1}
\end{figure*}

\begin{figure*}[h!]
  \centering
  \includegraphics[width=\linewidth]{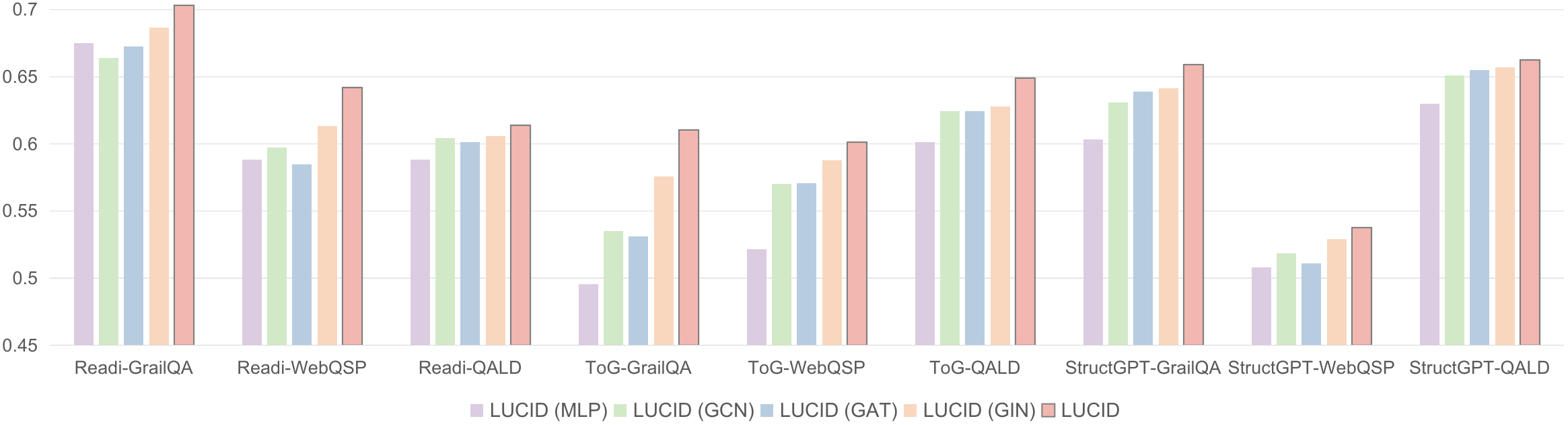}
  \caption{AVG performance of the method on three frameworks and three datasets when using different trained models.}
  \label{fig:ablation2}
\end{figure*}

\subsection{Main Results}

\textbf{Hallucination Detection Results.}
The performance of our proposed method (LUCID) and all baseline methods across three representative frameworks and three datasets is presented in Tables \ref{tab:main-readi}, \ref{tab:main-ToG}, and \ref{tab:main-StructGPT}. Table \ref{tab:main-readi} reports results using Readi, Table \ref{tab:main-ToG} uses ToG, and Table \ref{tab:main-StructGPT} employs StructGPT. 

Results show that LUCID exhibits superiority in the hallucination detection task, achieving state-of-the-art (SOTA) performance on all tested frameworks and datasets. 
On the AVG metric, it outperforms SelfCheckGPT by an average of 6.76\% and REDEEP (chunk) by an average of 5.48\%.
Furthermore, the model maintains high efficiency with an average inference time of 0.04 milliseconds per sample.

Compared to the best baseline on each dataset, LUCID achieves the largest average gain on Readi among the three frameworks (+2.81\%).This can be attributed to Readi’s reasoning-path editing mechanism, which iteratively refines retrieved KG paths and incorporates more valid triples into the subgraph. Consequently, Readi produces KG subgraphs with lower sparsity and richer local connectivity, as shown in Table \ref{tab:compare}. Such graph structures are particularly suitable for LUCID, whose message-passing mechanism benefits from denser local connectivity. Training on these graphs enables LUCID to better capture structural consistency and attention distribution patterns, leading to more reliable hallucination detection.
In contrast, ToG and StructGPT tend to retrieve sparser subgraphs, limiting the structural evidence available for detection and resulting in comparatively smaller (though still consistent) gains.

From the dataset perspective, LUCID achieves the most pronounced improvements on WebQSP (+2.57\% on average). WebQSP contains a higher proportion of relation-centric and single-/two-hop questions, where hallucinations often stem from incorrect relation selection despite surface-level semantic plausibility. LUCID is particularly effective in this setting, as it can suppress such structurally unsupported relation choices.
On QALD-10, which features diverse and complex query formulations, LUCID still delivers stable gains (+2.37\%), indicating robustness under higher linguistic and structural variability.

Across all settings, RAG-specific methods outperform general-purpose methods by a large margin (approximately +11.46\% on AVG). This highlights the importance of fine-grained evidence alignment in hallucination detection for KG reasoning. While existing methods focus on surface-level consistency of response and context, LUCID operates at a deeper level by aligning internal attention with graph structure.

\textbf{Utilize Hallucination Probabilities for QA Refinement.} In order to highlight that the probability of hallucinations detected by LUCID can be directly applied to improving the accuracy of QA, we evaluate their effectiveness as a signal for response refinement.

Specifically, we compare the following three QA strategies. The first two apply a single model to the entire dataset: Qwen2.5-7B and Qwen3-235B. Qwen2.5-7B is deployed locally on an NVIDIA RTX 4060 Ti (16GB) GPU with no monetary cost, whereas Qwen3-235B is accessed via API calls and incurs monetary cost.
The third strategy, denoted as \textit{Mixed}, performs selective refinement based on hallucination detections of LUCID. Responses are first generated by Qwen2.5-7B, and instances with high hallucination probabilities are refined using Qwen3-235B, while the remaining answers are retained.
For all strategies, we report exact-match (EM) accuracy and monetary cost (\$).

The results are shown in Table \ref{tab:QA}. The EM accuracy of Mixed is highly similar to that of the Qwen3-235B model across all datasets, with an average difference of 1.18\%. This indicates that refining certain outputs using the hallucination probabilities provided by LUCID can achieve performance close to that of refining all outputs. Besides, the cost of QA is substantially reduced in Mixed. Compared to Qwen3, Mixed saves an average of 55.4\% in monetary costs. These findings demonstrate that the hallucination detection mechanism of LUCID can significantly reduce costs while improving the accuracy of the model.

\subsection{Ablation Studies}\label{ablation}
We conduct several ablation studies to verify the importance of features and models used in Section \ref{Methodology}.  Here, we give nine groups of data from three datasets under three frameworks labeled as: Readi-GrailQA, ToG-GrailQA, StructGPT-GrailQA, Readi-WebQSP,  ToG-WebQSP, StructGPT-WebQSP, ToG-QALD, Readi-QALD, StructGPT-QALD. 

\subsubsection{Does the content of the feature matter?}
We modify the node and edge features to study the contribution of different feature components. Specifically, we evaluate the role of semantic similarity by removing it from edge features, and assess the importance of attention scores by replacing them with (1) hidden layer scores \cite{dai2022knowledge} (computed by averaging the scores for each node and edge token across all layers and heads and forming a final matrix), which encode the underlying features from the model's reasoning process, or (2) randomly initialized scores, while keeping feature weights unchanged.

We denote the complete method as "LUCID", the method that removes the similarity scores as "LUCID (w/o sim)", and the method that retains the similarity scores but replaces the attention matrices with random initialization matrices as "LUCID (w/o attn)". The one that replaces the attention matrices with the hidden layer scores is denoted as "LUCID (hidden)". The specific experimental results are shown in Figure \ref{fig:ablation1}.

Overall, LUCID consistently achieves the best performance across all datasets, demonstrating that combining LLM attention matrices with semantic similarity scores between KG relations and queries is crucial for hallucination detection. Removing semantic similarity leads to a moderate performance drop, indicating that attention matrices alone provide useful signals, while KG semantic information further enhances performance.

Replacing attention matrices with hidden layer scores yields only marginal gains over random attention, and both perform substantially worse than attention-based variants, confirming that LLM attention is the primary contributor, whereas hidden representations offer limited benefit.

\subsubsection{Does the choice of GNN model matter?}

To validate the effectiveness of the graph structure, we replace the GNN with an MLP for comparison. We also evaluate several alternative GNN models, including GCN \cite{kipf2017semi}, GAT \cite{velivckovic2017graph}, and GIN \cite{xu2018how}, to assess the advantage of the chosen GINE model. Since these models cannot natively handle edge features, we concatenate node features with their adjacent edge features as input. The training parameters of all other models are the same as those of the GINE model.

As shown in Figure \ref{fig:ablation2}, LUCID achieves the highest AVG across all experimental settings. Compared with LUCID (GIN), LUCID performs better by explicitly incorporating edge features into message passing, which better matches KG structure. LUCID (GCN) and LUCID (GAT) perform slightly worse, as they are less effective at modeling the full structural information of KGs.

Furthermore, LUCID (MLP) performs worse than all GNN-based variants, confirming the importance of graph structure. While MLPs process features in a flat manner, GNNs effectively capture the topological relationships in KGs through message passing, demonstrating the necessity of graph-aware modeling.

\subsection{Hyper-parameter Experiments}

\begin{figure}[h]
  \centering
  \includegraphics[width=\linewidth]{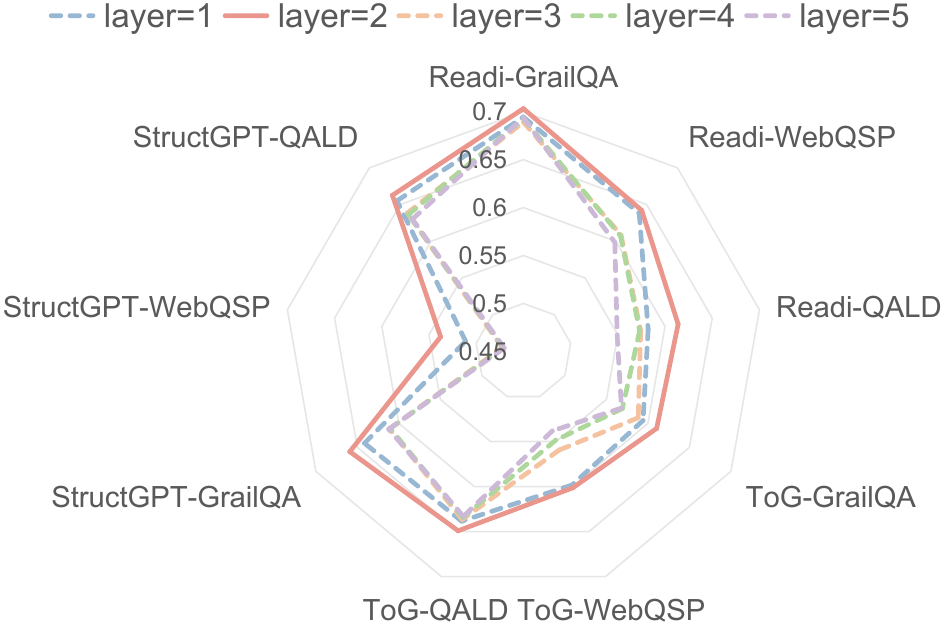}
  \caption{AVG comparison of methods on three frameworks and three datasets when using different numbers of GINE layers. The optimal result is shown as a solid line, while others are shown as dotted lines.}
  \label{fig:layer}
\end{figure}

\begin{figure}[h]
  \centering
  \includegraphics[width=\linewidth]{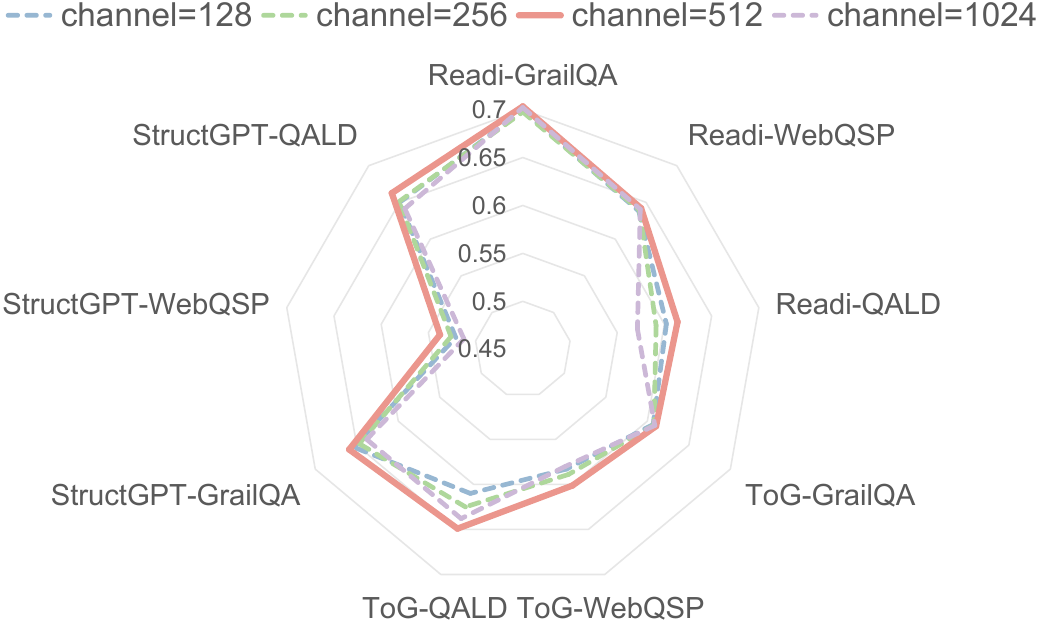}
  \caption{AVG comparison of methods on three frameworks and three datasets when using different numbers of GINE hidden channels. The optimal result is shown as a solid line, while others are shown as dotted lines.}
  \label{fig:channel}
\end{figure}

\begin{figure}[h]
  \centering
  \includegraphics[width=\linewidth]{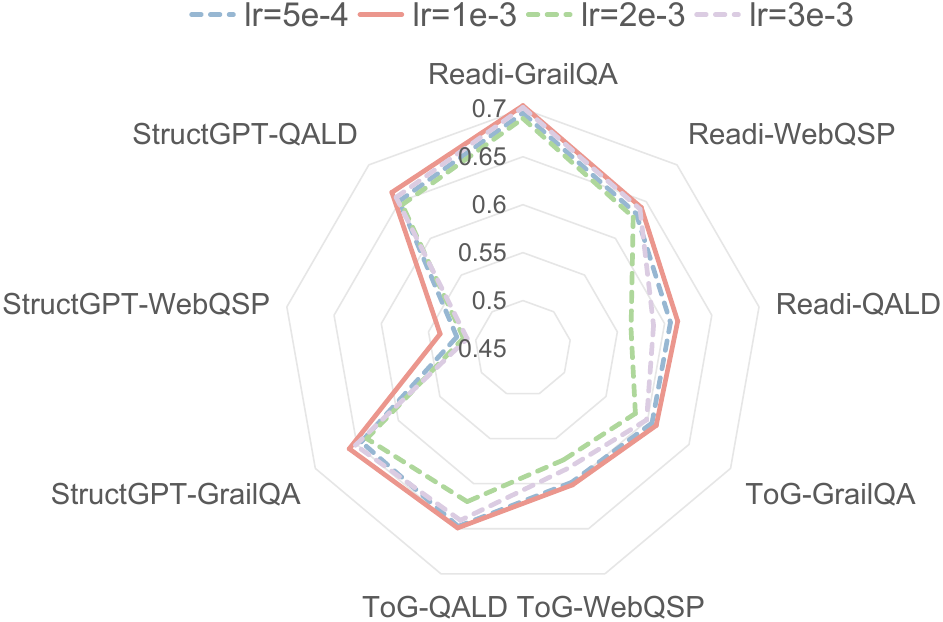}
  \caption{AVG comparison of methods on three frameworks and three datasets when using different learning rate. The optimal result is shown as a solid line, while others are shown as dotted lines.}
  \label{fig:lr}
\end{figure}

To further explore the impact of key hyper-parameters on the performance of the GINE model in hallucination detection, we conduct experiments on hyper-parameters such as the number of GINE layers, hidden layer channel size, and learning rate, under the same 9 sets of data as in Section \ref{ablation}. The specific results can be found in Figure \ref{fig:layer}, \ref{fig:channel} and \ref{fig:lr}.

\subsubsection{Does the number of layers matter?}
We evaluate the model with different numbers of GINE layers (1–5). As shown in Figure \ref{fig:layer}, the model achieves the best overall performance with 2 layers. Increasing the depth beyond 2 leads to marginal gains or performance degradation, and a sharp drop is observed with 4 or 5 layers, likely due to overfitting caused by excessive depth.

\subsubsection{Does the channel of hidden layers matter?}
We evaluate the model performance with different hidden layer channel sizes (128, 256, 512, 1024). Figure \ref{fig:channel} shows that 512 channels yield the best performance. Smaller channel sizes limit the model’s capacity to capture complex graph features, while excessively large channels degrade performance, likely due to overfitting.

\subsubsection{Does the learning rates matter?}
We test the model performance with different learning rates ($5\times10^{-4}$, $1\times10^{-3}$, $2\times10^{-3}$, $3\times10^{-3}$). As shown in Figure \ref{fig:lr}, a learning rate of $1\times10^{-3}$ performs best. Lower rates slow convergence, whereas higher rates lead to unstable optimization and reduced performance.

\section{Conclusion}
This paper addresses the challenges of hallucination detection in LLM-based KG reasoning frameworks by proposing LUCID, a novel hallucination detection method that integrates LLM internal state information,  KG semantic and structural information. 

For future work, we aim to further build upon the LUCID framework along several promising directions. First, we plan to explore fine-grained hallucination attribution to improve interpretability and error analysis. Second, we will extend evaluation settings to broader KG environments, including multilingual and cross-domain knowledge graphs, to assess the robustness and generalization of LUCID under diverse real-world conditions. These extensions will further advance the reliability of LLM-based KG reasoning systems for high-stakes applications such as decision support and automated knowledge discovery.

\bibliography{custom}

\appendix



\end{document}